\documentclass[journal]{IEEEtran}
\usepackage{cite}
\usepackage{amsmath,amssymb,amsfonts,amsthm}
\usepackage{algorithm}
\usepackage{algorithmic}
\usepackage[algo2e,linesnumbered,ruled]{algorithm2e}
\usepackage{graphicx}
\usepackage{textcomp}
\def\BibTeX{{\rm B\kern-.05em{\sc i\kern-.025em b}\kern-.08em
    T\kern-.1667em\lower.7ex\hbox{E}\kern-.125emX}}
\usepackage{float}
\usepackage{color}

\usepackage[utf8]{inputenc}
\usepackage{mathtools}  
\usepackage{nccmath}
\providecommand{\norm}[1]{\lVert#1\rVert}
\usepackage{dsfont}
\usepackage{caption}
\usepackage{subcaption}
\theoremstyle{plain}

\newcommand{\bx}{{\mathbf x}}

\newcommand{\bu}{{\mathbf u}}

\newcommand{\by}{{\mathbf y}}

\newtheorem*{proof}{Proof}
\begin{document}

\title{Supervised DKRC with Images for Offline System Identification}
\author{Alexander Krolicki and Pierre-Yves Lavertu
\thanks{
The research was conducted during the course ME8930 Deep Learning in Engineering Applications as the final project taught by Dr. Rahul Rai at the Clemson University International Center for Automotive Engineering (CU-ICAR). We would like to recognize and thank Wenjian Hao, Dr. Bowen Huang, Dr. Yiqiang Han, Dr. Umesh Vaidya and others from the DIRA lab for their combined contribution to our understanding of the foundational materials of this paper.}
\thanks{Alexander Krolicki, is a PhD student in the Department of Mechanical Engineering, Clemson University, Clemson, SC 29631 USA (e-mail:  akrolic@clemson.edu)}
\thanks{Pierre-Yves Lavertu, is a PhD student the Department of Material Science and Engineering, Clemson University, Clemson, SC 29631 USA (e-mail: plavert@clemson.edu).}}
 
\maketitle


\begin{abstract}
Koopman spectral theory has provided a new perspective in the field of dynamical systems in recent years. Modern dynamical systems are becoming increasingly non-linear and complex, and there is a need for a framework to model these systems in a compact and comprehensive representation for prediction and control. The central problem in applying Koopman theory to a system of interest is that the choice of finite-dimensional basis functions is typically done apriori, using expert knowledge of the systems dynamics. Our approach learns these basis functions using a supervised learning approach where a combination of autoencoders and deep neural networks learn the basis functions for any given system. We demonstrate this approach on a simple pendulum example in which we obtain a linear representation of the non-linear system and then predict the future state trajectories given some initial conditions. We also explore how changing the input representation of the dynamic systems time series data can impact the quality of learned basis functions. This alternative representation is compared to the traditional raw time series data approach to determine which method results in lower reconstruction and prediction error of the true non-linear dynamics of the system.
\end{abstract}


\section{Introduction}
Dynamical systems are systems for which a function defines the time dependency of a point in a finite dimensional space. There are plenty of examples of such systems from airplanes flight paths to motion of a liquid in a container. While some can be described by sets of non-linear equations representing the desired physics, this approach can sometimes be a complex challenge. Numerical modeling is an appealing approach used in simulation which can help represent very complex non-linear systems which often leads to  computationally intensive simulations. This becomes cumbersome in controlled dynamical systems where the controller feedback to the dynamical system is time sensitive and lagging information can have severe consequences. \\
Over the past 2 decades or so, a new perspective of dynamics capable of approaching the challenge posed by the complexity of certain dynamical systems has emerged: Data-driven modeling. These methods are capable of determining the spectrum of high-dimensional, non-linear dynamical systems. The specifics of these methods will be introduced in detail in the following section. Most popular current methods derive from Dynamic Mode Decomposition (DMD)\cite{schmid_2010}. This method has led to promising techniques and more recently deep learning has grabbed most of the research attention in this field of research. All derivative methods of DMD rely to some extent on the Koopman operator theory \cite{Koopman} also described later. DMD and deep learning approaches have already been used in many different areas with a substantial level of success and some of those examples are now presented. \\

\subsection{Data-driven modeling method relevance through examples:} 
There is a broad spectrum of applications that can benefit from the Data-driven modeling methods though not all dynamical systems are well suited. As mentioned already, one important aspect of the deep neural network approach to learn the basis functions is that it is heavily dependent on access to existing data or ways to generate the required data. A good example and relevant application is environment control systems in buildings, where large amounts of sensor information is captured and even virtual building simulation can be used to generate datasets. In this case, using Koopman operators facilitates the comparison of complex data while also simplifying the system representation. In Eisenhower et al. 2016\cite{Eisenhower}, the spectral decomposition approach is used to analyze building system data by accelerating the comparison between models and data as well as drawing conclusions about the sensor functions. The authors show that using the spectral decomposition can help understand the changes in the different parts of a building and help highlight poor control performance. 

A similar approach can also be applied to power grid systems to help evaluate the response to continuously changing local demand. In this case, the challenge is in the size of the system and the amount of heterogeneous dynamic sub-systems like power plants, transmission lines and other renewable energy systems. The objective in using the Koopman operator technique here is to help identify key dynamic phenomena to help understand cascading power outages. Susuki et al. 2017\cite{Susuki} successfully used Koopman Mode Decomposition (KMD) to enable direct computation from data without describing the complex underlying system. 

More examples can be found in Xiao et al. 2020 \cite{Xiao} (Vehicle motion planning and control algorithms), Broad et al. 2019 \cite{Broad} (Human-Machine Systems to help users accomplish tasks), Ling et al. 2019 \cite{Ling} (Intelligent Transport Systems (ITS) to help with the reduction of fuel consumption), and Fonzi et al. 2020 \cite{Fonzi} (nonlinear dynamics fundamentals to the morphing airborne wind energy (AWE) aerostructures).

\subsection{Existing works:} 
Advances in research from both the controls and computer science communities has contributed to much of the underlying mechanisms and tools discussed in this paper. Our proposed approach is a direct improvement of the original DKRC paper \cite{Han}, where we introduce the autoencoder to learn the basis functions. Exisitng works have utilized the same approach for just processing in the raw data alone \cite{Lusch_2018}. Other works have already considered integrating images into this framework such as DKRC-I \cite{DBLP:conf/crv/LaferriereLDFP21} and CKNet \cite{xiao2021cknet}. Our work looks at combining both the raw time series data and a spectrogram image which is derived only from the raw time series data as an additional input feature to the autoencoder network. Closer examination of the pros and cons between these existing works and how the underlying DKRC code could be restructured are outside of the scope of this paper but will be explored in future research.

\section{Preliminaries and Notations}\label{section_prelim}

The following preliminaries are cited from \cite{Huang}. In order to define the properties of the Koopman operator, lets take a discrete time dynamical system $x_{t+1}=\mathrm{T}(x_t)$ where $\mathrm{T}:\mathrm{X} \subset \mathbb{R}^n \rightarrow \mathrm{X}$. We also denote $\mathcal{B}\left(X\right)$ the Borel-$\sigma$ algebra on $\mathrm{X}$, $\mathcal{M}(\mathrm{X})$ a vector space of bounded complex valued measure on $\mathrm{X}$, and $\mathcal{F}$ the space of complex valued functions from $ X \rightarrow \mathbb{C}$. Associated with this discrete time dynamical system is the linear operator, $\mathbb{U}$, called the Koopman operator. The Koopman operator is an infinite-dimensional linear operator defined on the space of functions as follows:

\begin{equation}\label{eqn.koop}
\begin{gathered}
\left[\mathbb{U}\varphi\right]\left(x\right)=\varphi(T(x))
\end{gathered}
\end{equation}

Where the observable function $\varphi$ is mapped forward in time by the Koopman operator. The spectrum (eigenvalues and eigenfunctions) of the Koopman operator satisfy the following relationship

\begin{equation}\label{eqn.koop_eig}
\begin{gathered}
\left[\mathbb{U}_t\phi_\lambda\right]\left(x\right)=e^{\lambda t}\phi_\lambda(x)
\end{gathered}
\end{equation}

Where $\phi_\lambda$ is an eigenfunction and $\lambda\in\mathbb{C}$ is the associated eigenvalue. The eigenfunctions of the Koopman operator can be used as coordinates for the linear representation of nonlinear systems. The relationship between the spectrum of the Koopman system and stability is explored in \cite{Mauroy2013ASO}.\\

Dynamic Mode Decomposition (DMD) is a computational algorithm for approximating the spectrum of the Koopman operator \cite{schmid_2010}. Extended Dynamic Mode Decomposition (EDMD) is more accurate in approximating the spectrum of the Koopman operator for both linear and non-linear dynamical systems \cite{Williams_2015}. The formulation of EDMD is as follows.

\begin{equation}\label{eqn.data}
\begin{gathered}
\bar{X}=[x_1,x_2,\ldots,x_M] \\
\bar{Y}=[y_1,y_2,\ldots,y_M]
\end{gathered}
\end{equation}

Where $x_i\in X$and $y_i\in X$. The set $\bar{Y}$ is the time shifted time series data such that $y_i=T(x_i)$. Let $\mathcal{D}=\left\{\psi_1,\ \psi_2,\ \ldots,\ \psi_N\right\}$ be the set of dictionary functions or observables where $\psi_i\in\ L_2(X,\mathcal{B},\mu=\mathcal{G})$. Here $\mu$ is a positive invariant measure, but not necessarily an invariant measure of $T$. Let $\mathcal{G}_\mathcal{D}$ denote the span of $\mathcal{D}$ such that $\mathcal{G}_\mathcal{D}\subset\mathcal{G}$. The choice of dictionary functions are critical to the accuracy of the approximated eigenfunctions of the Koopman operator. Define the vector valued function $\Psi : X \rightarrow \mathbb{C}^N$

\begin{equation}\label{eqn.dict}
\begin{gathered}
\Psi\left(x\right):=[\psi_1(x),\psi_2(x),…,\psi_N(x)]^T
\end{gathered}
\end{equation}

$\Psi$ is the mapping from physical space to feature space. Any function $\phi,\hat{\phi}\in\mathcal{G}_\mathcal{D}$  with some set of coefficients $a, \hat{a}\in\mathbb{C}^N$ can be expressed as,

\begin{equation}\label{eqn.psia}
\begin{gathered}
\phi=\sum_{k=1}^{N}{a_k\psi_k=\Psi^Ta}
\end{gathered}
\end{equation}

\begin{equation}\label{eqn.psiahat}
\begin{gathered}
\hat{\phi}=\sum_{k=1}^{N}{{\hat{a}}_k\psi_k=\Psi^T\hat{a}}
\end{gathered}
\end{equation}

Let the future observable $\hat{\phi}$ be related to current observable $\phi$ by the Koopman operator in the feature space. We can write

\begin{equation}\label{eqn.psir}
\begin{gathered}
\hat{\phi}\left(x\right)=\left[\mathbb{U}\phi\right]\left(x\right)+r
\end{gathered}
\end{equation}

Where $r\in\mathcal{G}$ is a residual function that appears because $\mathcal{G}_\mathcal{D}$ is not necessarily invariant to the action of the Koopman operator. To find the optimal mapping which can minimize this residual such that we can obtain

\begin{equation}\label{eqn.psinor}
\begin{gathered}
\hat{\phi}\left(x\right)=\left[\mathbb{U}\phi\right]\left(x\right)
\end{gathered}
\end{equation}

We need to choose a basis function such that the span of $\mathcal{D}$ contains the minimum least squares solution to the following. 

\begin{equation}\label{eqn.G}
\begin{gathered}
G=\frac{1}{M}\sum_{m=1}^{M}{\Psi\left(x_m\right)\Psi\left(x_m\right)^T}
\end{gathered}
\end{equation}

\begin{equation}\label{eqn.A}
\begin{gathered}
A=\frac{1}{M}\sum_{m=1}^{M}{\Psi\left(x_m\right)\Psi\left(y_m\right)^T}
\end{gathered}
\end{equation}

\begin{equation}\label{eqn.minopt}
\begin{gathered}
\min_{x}\norm{GK-A}_F
\end{gathered}
\end{equation}

The symbol $\norm{\cdot}_F$ denotes the Frobenius norm of a matrix. The explicit least squares solution to this optimization problem is given as,

\begin{equation}\label{eqn.Kedmd}
\begin{gathered}
K_{EDMD}=G^\dag A
\end{gathered}
\end{equation}

Where $G^\dag$ the Moore-Penrose pseudoinverse of matrix G. Therefore, assuming the dictionary of basis functions $\Psi$ spans the subspace $L_2(X,\mathcal{B},\mu)$ then we can say that the leading eigenfunctions of the Koopman operator and the associated eigenvalues given by the EDMD approximation can be computed. The right eigenvectors of $K$ generate the approximation of the eigenfunctions, given by,

\begin{equation}\label{eqn.Keigfunc}
\begin{gathered}
\phi_j=\Psi^Tv_j
\end{gathered}
\end{equation}

Where $v_j$ is the $j$-th right eigenvector of $K$, $\phi_j$ is the eigenfunction approximation of the Koopman operator associated with the $j$-th eigenvalue. 

DMD approximates the Koopman operator with a specific choice of dictionary functions chosen to be the unit vectors $e_i\in\mathbb{R}^n$ of the lifted vector space.

\begin{equation}\label{eqn.unitbasis}
\begin{gathered}
e_1=\left[\begin{matrix}1\\0\\\vdots\\\end{matrix}\right],\ e_2=\left[\begin{matrix}0\\1\\\vdots\\\end{matrix}\right],\ \ldots
\end{gathered}
\end{equation}

\begin{equation}\label{eqn.unitset}
\begin{gathered}
\mathcal{D}=\{e_1^T,\ldots,e_N^T\}
\end{gathered}
\end{equation}
such that the least square solution to the DMD Koopman operator can be solved directly from the data.

\begin{equation}\label{eqn.Kdmd}
\begin{gathered}
K_{DMD}=\bar{Y}\ {\bar{X}}^\dag
\end{gathered}
\end{equation}

In this paper, the EDMD formulation is solved in the lifted space for a controlled dynamical system as formulated in \cite{Korda_2018}. therefore we can first redefine our datasets as,

\begin{equation}\label{eqn.dataxlift}
\begin{gathered}
X_{lift}=[\psi\left(x_1\right),\ \ldots,\ \psi\left(x_k\right)] \\ 
Y_{lift}=\left[\psi\left(y_1\right),\ \ldots,\ \psi\left(y_k\right)\right] \\ 
U=[u_1,\ldots,u_k]
\end{gathered}
\end{equation}

We want to obtain the lifted linear dynamical system

\begin{equation}\label{eqn.linearlifted}
\begin{gathered}
Y_{lift}=AX_{lift}-BU \\ 
\hat{x}=CX_{lift}
\end{gathered}
\end{equation}

The least squares solution for the lifted linear state space system is solved using \eqref{eqn.ab},\eqref{eqn.c}. Where $N$ is the lifting dimension, $m$ is the number of data snapshots, and the dimensions are $A_{Nx(1:N)}$, $B_{Nx(N+1:m)}$, $C_{nxN}$.

\begin{equation}\label{eqn.ab}
\begin{gathered}
\left[A,B\right]=\left[Y_{lift}\left[\begin{matrix}X_{lift}\\U\\\end{matrix}\right]^T\right]{\left[\left[\begin{matrix}X_{lift}\\U\\\end{matrix}\right]\left[\begin{matrix}X_{lift}\\U\\\end{matrix}\right]^T\right]^{-1}}
\end{gathered}
\end{equation}

\begin{equation}\label{eqn.c}
\begin{gathered}
C=XX_{lift}^\dag
\end{gathered}
\end{equation}

This solution leads to a better fit of the lifted linear dynamical system since the solution is obtained in the the lifted space. 

\section{Algorithm}

Our implementation consists of several sub-processes which makeup the overall pipeline for determining the final linear lifted representation of the non-linear system. Provided time series data which contains measured state variables and control inputs, we first perform data pre-processing. During pre-processing, a Mel spectrogram image is created by sampling the available time series dataset at a frequency less than the data was measured. In this case, we lose some of the time resolution of the data, so in cases where high frequency sampling is not available, this approach may not be as data efficient. Most modern engineering systems do have high sampling rates, so this is not a major concern but rather something to consider when applying it to the system of interest. Once the spectrogram images are created, a convolutional autoencoder is trained to minimize the reconstruct error of the input image. At the output of the encoder in the convolutional autoencoder network, there is a bottleneck layer which captures the latent features of the images. These latent features are then combined together with the raw time series data as inputs to a fully connected autoencoder network. Similarly, the autoencoder trains to minimize the reconstruction error of the enriched input representation. The bottleneck layer of this network is equal to the lifting dimension, and once the mean absolute error between the input and output of the network is minimized, we can remove the decoder. Now we can use the encoder as the lifting basis function which we call the lifting DNN. Finally, the lifting DNN maps latent image data and raw time series data to the lifted state space.  In the end, we want to find a linear state space model which minimizes reconstruction error as well as prediction error compared to the real non-linear system. Another criteria for this system is that it must be controllable. Once these 2 criteria are satisfied, the linear system is evaluated to see how well it can track the state trajectories given some initial conditions and control inputs. Further analysis on the controllability and stability of these systems is beyond the scope of this paper, as we are only focused on the model predictive element of the system identification problem.

\section{Convolutional Autoencoder}
\begin{figure}[H]
\centering
\includegraphics[width=0.5\textwidth]{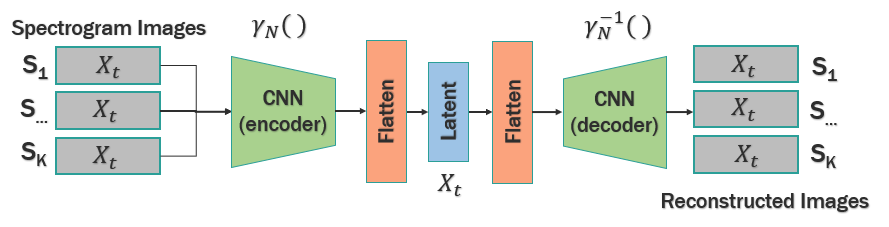}
\caption{Convolutional autoencoder network mapping from pixel space to a latent vector space then back to pixel space.}
\label{fig:cae}
\end{figure}

\begin{figure}[H]
\centering
\includegraphics[width=0.5\textwidth]{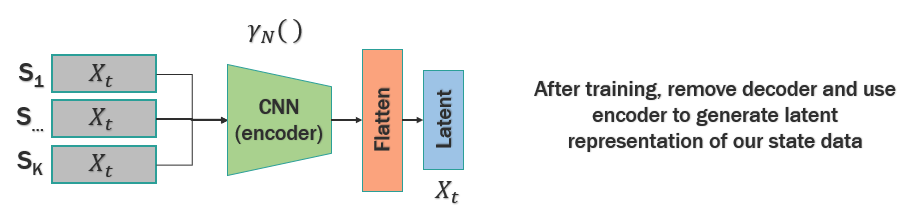}
\caption{Encoder network used to generate latent labels for each spectrogram image}
\label{fig:ce}
\end{figure}

The purpose of the convolutional autoencoder network (CAE) shown in figure \ref{fig:cae} is to learn the encoding from pixel space to the latent vector space and then decoding back to the pixel space minimizing the reconstruction error of the spectogram input image. The loss function used to update the weights and biases of the network is simply the mean square error. The encoder portion is composed of two convolution layers with LeakyReLU activation functions and including 'same' padding. The decoder is also using LeakyReLU activation function with the exception of the last activation function, which is Sigmoid. The main objective is to extract the latent representation of the state data as depicted in figure \ref{fig:ce}.

After training we remove the decoder from the CAE. The encoder section shown in figure \ref{fig:ce} generates the latent labels from the input images which are then concatenated with the raw state data. 

\begin{algorithm2e}
\textbf{Input: } \\
dataset containing all data up to time $T$ $D = \{\bx_i,\by_i,\bu_i, q_i\}_{i=0}^{T}$\\
training loss from previous training iterations $L_{tot}$\\
\textbf{Output: } \\
Linear Lifted System ${X_{t+1}}_{lift}=A{X_t}_{lift}+BU_t$\\
lifting basis function DNN $\psi_{\theta}(\bx)$\\
\SetAlgoLined
\rule{80mm}{0.3mm}\\[-1pt]
initialize DNN weights $\theta$\\
define number of training epochs $n_t$ \\
let $\psi_{\theta}$ be the lifting network $\psi_\theta:x_{nx1}\rightarrow\ X_{Nx1}$ \\
let $\psi_{\theta}^{-1}$ be the decoding network $\psi_\theta^{-1}:\ X_{Nx1}\rightarrow\ x_{nx1}$ \\
given the lifted system ${\psi_\theta(x_{t+1})=A\psi}_\theta\left(x_t\right)+Bu_t$ \\
which can also be written as $X_{t+1}=AX_t+Bu_t$\\
\ \\

\For{$n_t$}{
    generating lifted states using $\psi_{\theta}$\\
    ${X_t} = \psi_{\theta}(\bx)$\\
    ${Y_{t}} = \psi_{\theta}(\by)$\\
    computing the bi-linear system\\
    $\left[A,B\right]=\left[Y_{t}\left[\begin{matrix}{X_t}\\U_t\\\end{matrix}\right]^T\right]\left[\left[\begin{matrix}{X_t}\\U_t\\\end{matrix}\right]\left[\begin{matrix}{X_t}\\U_t\\\end{matrix}\right]^T\right]^{-1}$\\
    compute linearization loss \\
    $L_1=X_{t+1}-[A{X_t}+BU_t]$\\
    compute controllability loss \\
    $L_2=N_{lift}-rank(ctrb\left(A,B\right))$\\
    compute encoder loss \\ 
    $L_3=\psi_\theta^{-1}\left(X_{t+1}\right)-x_{t+1}$\\
    compute total loss \\
    $L_{tot} = L_1 + L_2 + L_3$\\
    \uIf{$L_{tot} < min(L_{tot})$ and $L_2=0$}{
    solve \eqref{eqn.ab} and \eqref{eqn.c} and store $A$,$B$,$C$ and $\theta_{final}$ \\
    }
    \Else{
    backpropagate loss and update $\psi_{\theta}^{-1}$ and $\psi_{\theta}$  $\theta_{old} \rightarrow \theta_{new}$ \;
    }
}

\caption{Unsupervised DKRC}\label{alg1}
\end{algorithm2e} 

\begin{algorithm2e}
\textbf{Input: } \\
dataset containing all data up to time $T$ $D = \{\bx_i,\by_i,\bu_i, q_i\}_{i=0}^{T}$\\
\textbf{Output: } \\
Linear Lifted System ${X_{t+1}}_{lift}=A{X_t}_{lift}+BU_t$\\
lifting basis function DNN $\psi_{\theta}(\bx)$\\
\SetAlgoLined
\rule{80mm}{0.3mm}\\[-1pt]
initialize DNN weights $\theta$\\
define number of training epochs $n_t$ \\
define the accuracy tolerance $\epsilon$ \\
let $\psi_{\theta}$ be the lifting network $\psi_\theta:x_{nx1}\rightarrow\ X_{Nx1}$ \\
given the lifted system ${\psi_\theta(x_{t+1})=A\psi}_\theta\left(x_t\right)+Bu_t$ \\
which can also be written as $X_{t+1}=AX_t+Bu_t$\\
lifting dimension $N=n+1$
\ \\

\For{$n_t$}{
    train CAE $\gamma_{\theta}$ and $\gamma_{\theta}^{-1}$ \\
    train AE $\psi_{\theta}(x)$ and $\psi_{\theta}^{-1}(x)$\\
    generate spectrogram latent states using $\gamma_{\theta}$\\
    generate lifted states using $\psi_{\theta}$\\
    ${X_t} = \psi_{\theta}(\bx)$\\
    ${Y_{t}} = \psi_{\theta}(\by)$\\
    computing the bi-linear system\\
    $\left[A,B\right]=\left[Y_{t}\left[\begin{matrix}{X_t}\\U_t\\\end{matrix}\right]^T\right]\left[\left[\begin{matrix}{X_t}\\U_t\\\end{matrix}\right]\left[\begin{matrix}{X_t}\\U_t\\\end{matrix}\right]^T\right]^{-1}$\\
    compute linearization heuristic \\
    $h_1=X_{t+1}-[A{X_t}+BU_t]$\\
    compute controllability criteria \\
    $h_2=N_{lift}-rank(ctrb\left(A,B\right))$\\
    compute total loss \\
    $h_{tot} = h_1$\\
    \uIf{$h_{tot} < \epsilon$ and $h_2=0$}{
    solve \eqref{eqn.ab} and \eqref{eqn.c} and store $A$,$B$,$C$ and $\theta_{final}$ \\
    }
    \Else{
    increment lifting dimension $N=N+1$\;
    }
}

\caption{Supervised DKRC}\label{alg2}
\end{algorithm2e} 

\section{Autoencoder}
\begin{figure}[H]
\centering
\includegraphics[width=0.5\textwidth]{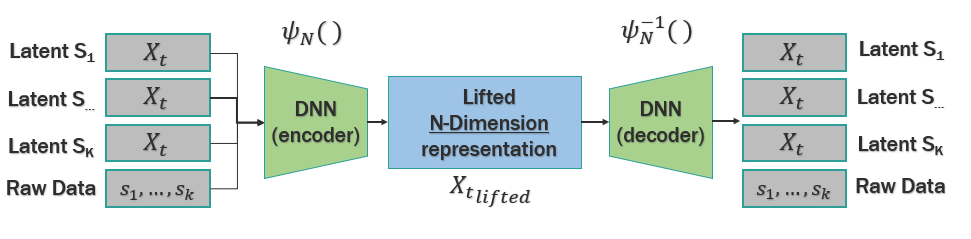}
\caption{Fully connected autoencoder network}
\label{fig:ae}
\end{figure}

The main purpose of the autoencoder network (AE) is to learn the lifted $N$-th dimension latent space mapping shown in Figure \ref{fig:ae}. The AE is the core of the supervised data-driven approach for the learning of the basis functions used to compute the Koopman operator.\\  

\begin{figure}[H]
\centering
\includegraphics[width=0.5\textwidth]{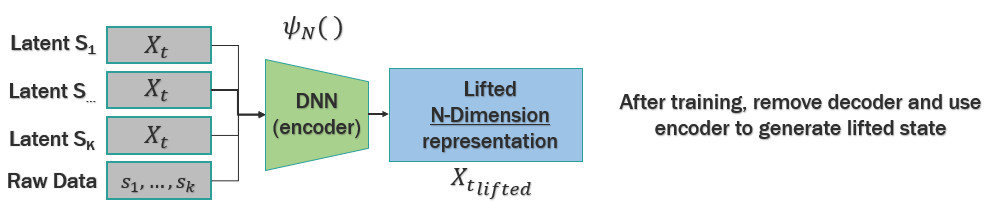}
\caption{Fully connected encoder network (lifting basis function) mapping latent image data and raw time series data to the lifted N-dimensional state.}
\label{fig:e}
\end{figure}

The encoder portion is composed of a single dense layer using the tanh activation function. Following training of the AE, the decoder is removed and the lifted states can be predicted as shown in Figure \ref{fig:e}.

\begin{figure}[H]
\centering
\includegraphics[width=0.5\textwidth]{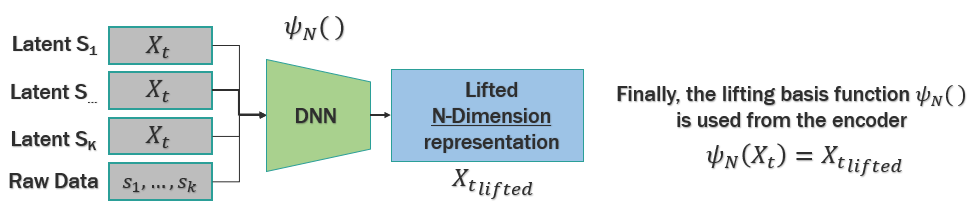}
\caption{Deep Neural Network that maps the latent image representations and raw data to the lifted state}
\label{fig:lift}
\end{figure}

Once the lifting basis functions \begin{math} \psi_{\theta}\left(x\right) \end{math} have been learned, we can predict the lifted state for \begin{math} {X}_{t_{lift}} \end{math} as shown in figure \ref{fig:lift}. 

In our case, we are dealing with forced dynamical systems, which require a different set of equations to compute the lifted linear representation of the system. The derivation of these formulas used in this section are beyond the scope of this text, but are cited for reference \cite{Korda_2018}.

In order to obtain the state space representation of our lifted linear forced dynamical system, we compute $A$, $B$, and $C$ matrices where $D$ is equal to zero. 

\begin{equation}\label{eqn.AB}
\begin{gathered}
\left[A,B\right]=\left[X_{t+1_{lift}}\left[\begin{matrix}{X_t}_{lift}\\U_t\\\end{matrix}\right]^T\right]\left[\left[\begin{matrix}{X_t}_{lift}\\U_t\\\end{matrix}\right]\left[\begin{matrix}{X_t}_{lift}\\U_t\\\end{matrix}\right]^T\right]^{-1}
\end{gathered}
\end{equation}

\begin{equation}\label{eqn.C}
\begin{gathered}
C=X_t\left[{X_t}_{lift}\right]^T
\end{gathered}
\end{equation}

\begin{equation}\label{eqn.D}
\begin{gathered}
D=0
\end{gathered}
\end{equation}

In order to evaluate the accuracy of the lifted system, we compute the heuristic term $h_1$ using equation \ref{eqn.L1} which compares the lifted state at the next time step to the linear dynamic systems output which should result in zero if the approximation is accurate. This is in contrast to the original unsupervised DKRC algorithm (shown in algorithm 1) proposed in \cite{Han}, where the error is computed as a loss which is used to update the networks weights. 

\begin{equation}\label{eqn.L1}
\begin{gathered}
{h_1=\ X}_{t+1_{lift}}-[A{X_t}_{lifted}+BU_t]
\end{gathered}
\end{equation}

The controllability matrix $Q$ is defined in equation \ref{eqn.ctrb}. The rank function determines the number of linearly independent columns in $Q$. By definition, a linear dynamical system is controllable if the rank of the controllability matrix is equal to the order of the system which in this case is equal to the lifted dimension $N$. 

\begin{equation}\label{eqn.ctrb}
\begin{gathered}
Q=\left[\begin{matrix}B&AB&A^2B\\\end{matrix}\ \ \begin{matrix}\ldots&A^{N-1}B\\\end{matrix}\right]
\end{gathered}
\end{equation}

Hence, we define our second loss function $L_2$ to be equal to the difference between the rank of the controllability matrix and the order of the system $N$.

\begin{equation}\label{eqn.L2}
\begin{gathered}
h_2=N_{lift}-rank(Q)
\end{gathered}
\end{equation}

So finally, the total heuristic value $h$ is computed in equation \ref{eqn.losstot} as equal to the lifted prediction error heuristic $h_1$. The controllability criteria $h_2$ is not included in the total heuristic value since we only want to admit solutions which are controllable such that $h_2=0$. In order for an identified linear system to be admissible, it must have a total loss less than some constant $\epsilon$. Using the combination of heuristic measures and admissibility criteria, the final identified system will be controllable and accurate with in the specified tolerance. If a solution cannot be found at the lifting dimension $N=n+1$ then the AE networks latent space is increased to dimension $N=N+1$ for as many times as it takes to obtain a admissible solution. 

\begin{equation}\label{eqn.losstot}
\begin{gathered}
h_{tot} = h_1<\epsilon
\end{gathered}
\end{equation}

\begin{equation}\label{eqn.control}
\begin{gathered}
rank(Q) = N
\end{gathered}
\end{equation}
The hypothesis here is that since we are approximating an infinite dimensional operator and deriving our lifted linear system from that approximation of the operator, then the larger dimension $N$ we choose should converge to the exact solution in the limit. In this case $\epsilon$ is equal to zero and the original nonlinear dynamics maps perfectly lifted linear state space system. Therefore, we could say that as $N$ increases towards infinity, $h_1$ goes to zero. 
\begin{equation}\label{eqn.hyp}
\begin{gathered}
\lim_{N\rightarrow\infty} X_{t+1_{lift}}-[A{X_t}_{lifted}+BU_t]=0
\end{gathered}
\end{equation}
This is why if the solution is not accurate or the controllability criteria is not met, we increase the size of N incrementally. The proposed algorithm is shown in algorithm 2.

\section{Simulation Results}\label{section_main}
In this study, the usage of spectogram images with raw time series data is compared to only raw time series data as the input in order to determine if the quality of the learned basis functions can be improved with additional features included in the input (which are derived from the time series data). The main objective being to increase the quality of the learned basis functions such that we can accurately predict the dynamics for the design of a controller (MPC, LQR, ect). From this, we identified three metrics to evaluate the success of our model. The first metric is the complexity of the model setup in terms of data gathering and preparation, as well as the decoding of the model outputs for interpretation. A second metric considered is the dimension of the A and B matrices. Low dimension A and B matrices are targeted as it would require a lower computational effort in computing control signals. The last metric in this study is the average state error of a trajectory of states $\theta$ and $\dot{\theta}$. We used Python to implement the proposed algorithms with TensorFlow being the foundation for all neural networks in the architecture. The OpenAI gym 'Pendulum-V0' \cite{brockman2016openai} was modified to step forward at a frequency of 1,000 Hz in order to increase the sampling frequency for the spectrogram images time resolution to match well with the frequency resolution.

\subsection{Metric 1: Complexity of the model setup}
If we evaluate the model structure of different publications like Han et al. \cite{Han} and the example presented in Brunton et al. \cite{Brunton}, they use DNN structures only, which do not require the extra step of learning the latent space representation of the raw data. Leaving aside the autoencoders themselves and focusing on the data pre-processing step, the simple fact that the proposed model uses spectrogram images means that the data undergoes an extra transformation. The time series data conversion to image is a direct one but as the training progresses to the networks and reaches the first DNN encoder, it then gets merged with the initial raw data which increases the complexity of the architecture. Altogether, having this extra input feature of the spectrogram image does induce greater algorithmic and time complexity into solving this problem. Therefore, in evaluating the first criteria, the proposed model brings extra complexity with the addition of the CNN in comparison to the other models presented in literature \cite{Han} \cite{Brunton}. The hope is that this added complexity is compensated for in the fidelity of the identified system.

\subsection{Metric 2: Size of the A and B matrices}
\begin{figure}[htbp]
\centering
\includegraphics[width=2.0in]{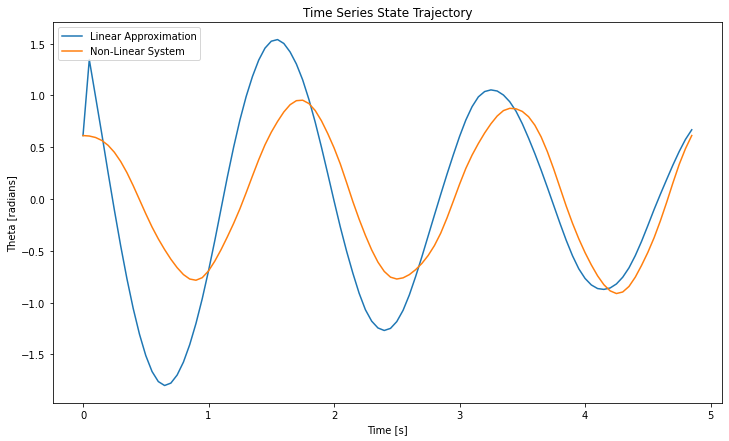}
\hspace{0.1cm}
\includegraphics[width=2.0in]{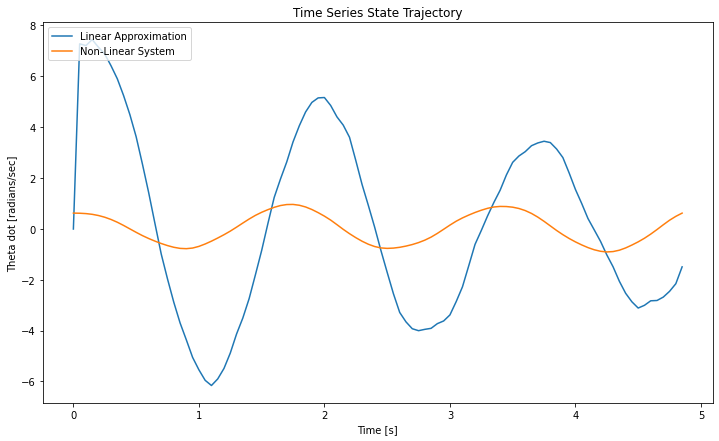}
\caption{Raw data + Latent image data compared to non-linear system lifting dimension = 12}
\label{fig:latlift12}
\end{figure}

\begin{figure}[htbp]
\centering
\includegraphics[width=2.0in]{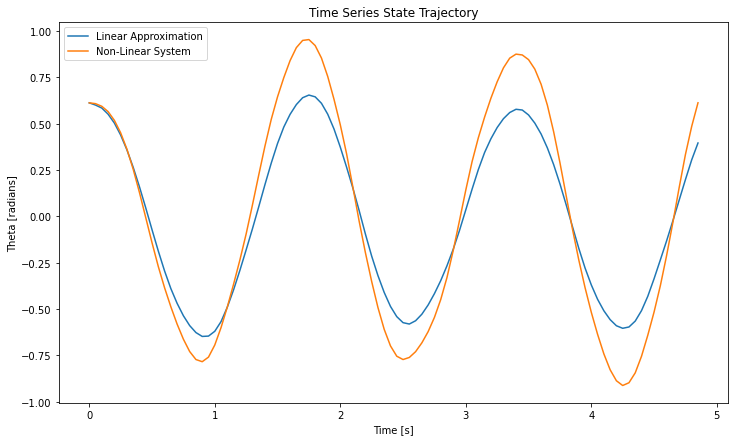}
\hspace{0.1cm}
\includegraphics[width=2.0in]{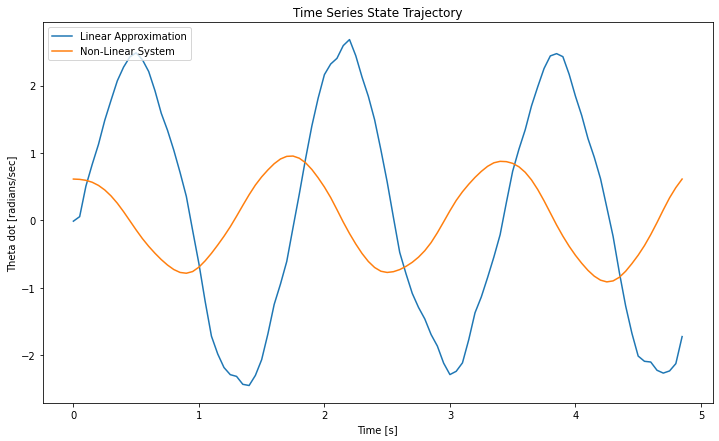}
\caption{Raw data only compared to non-linear system lifting dimension = 12}
\label{fig:rawlift12}
\end{figure}

Controlling the size of both A and B matrices enables a direct comparison for both the proposed approach and the model using the raw data only. The comparison here is done in terms of accuracy of the prediction. As the goal is to head toward a controller, the better model would be able to achieve good accuracy for an extended period of time. 
The maximum lifted dimension used in this study was \begin{math} N=12 \end{math} and it was also the lifting dimension giving the best prediction in both cases. In figure \ref{fig:latlift12}, it is possible to see that the utilization of the latent images generates poor initial predictions. With the raw data only in figure \ref{fig:rawlift12} initial predictions are good while prediction after 3 seconds are less accurate. The results shown in Table \ref{table:1} support the hypothesis of equation \eqref{eqn.hyp} in that the accuracy increases with higher dimension approximations.
In both cases, the angular velocity $\dot{\theta}$ is not well predicted. 

\begin{table}[H]
\centering
\begin{tabular}{ |p{3cm}||p{1cm}|p{1cm}|p{1cm}|p{1cm}| }
 \hline
 \multicolumn{5}{|c|}{Average State Trajectory Prediction Mean Absolute Error} \\
 \hline
 Lifting Dimensions & $\theta$* & $\dot{\theta}$* & $\theta$** & $\dot{\theta}$**\\
 \hline
 3rd Dimension & 0.515  & 1.95  & 0.30  & 1.18\\
 5th Dimension & 0.517  & 2.04  & 0.22  & 0.82\\
 12th Dimension & 0.456  & 1.84  & 0.14  &  0.54\\
 \hline
\end{tabular}
\caption{*Raw Data + Latent Image Data, **Raw Data Only}
\label{table:1}
\end{table}

\subsection{Metric 3: Average State Trajectory Prediction}
For this last criteria, a comparison of the mean absolute error over average state trajectory prediction is presented in Table \ref{table:1}. The results show an improvement for the raw data model only between the 3rd and 12th lifting dimension. The improvement trend only starts after the 5th dimension in the case where raw and latent image data is used for the basis function identification. 

\section{Discussion}\label{secction_simulation}
\begin{figure}[htbp]
\centering
\includegraphics[width=3.0in]{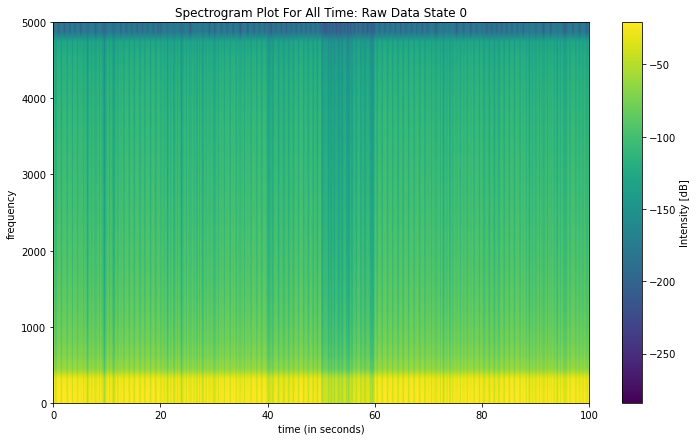}
\caption{Periodicity of pendulum Spectrogram image for all training time}
\label{fig:specbad}
\end{figure}

\begin{figure}[htbp]
\centering
\includegraphics[width=3.0in]{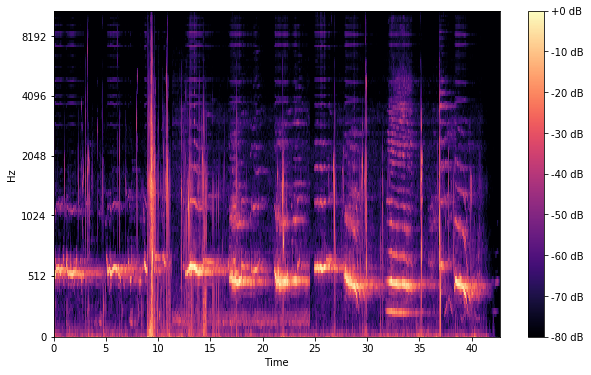}
\caption{Spectrogram Image from acoustic data}
\label{fig:specgood}
\end{figure}

One potential downfall of the simple pendulum system is that it is almost too simple of a non-linear system for the spectrogram image to bear any meaningful transient frequency information. To highlight this, we show our spectrogram image from the pendulum in figure \ref{fig:specbad}. It is clear that there is little difference distinguishing the trajectory of the dynamics of the measured state since it is repeatedly cycling through the same set of fixed values. A more complex example of a nonlinear system could be the measurement of acoustics such as what is shown in figure \ref{fig:specgood}. Visually it is clear that over time the state is evolving in a unique and identifiable manner, making the content of the image more useful for determining the trajectory of the audio or state. Ideally we would like to try out our approach on a more complex nonlinear system, especially a real system with noise and uncertainty to truly measure the robustness of the approach in performing accurate system identification.

\section{Conclusion}\label{section_conclusion}
The investigation around the use of spectogram images generated from a non-linear dynamical system showed that the approach can lead to accurate predictions, though it does not seem to be achieved as quickly as when only using raw data as the input. The current results provided by this work don't show conclusive evidence as to the benefit of the spectogram images in the quality increase in basis function identification at for low lifting dimensions. \\

Here, it is worth mentioning that only a single Neural Network format was evaluated and it would be worthwhile to evaluate different activation functions and layer structure in both the CAE and AE sections of the model. Also, the current model implementation uses simple CAE/AE networks and the usage of a variational autoencoders (VAE) networks should be considered. Existing work has shown that the use of VAEs and convolutional VAEs do a good job of reducing sample complexity in unexplored regions of the state space during sampling. While the spectogram images provided an enriched set of data, its accurate reconstruction through the initial autoencoder is crucial and could become a critical improvement leading to the actual desired outcome: An increased quality of the learned basis functions for more accurate prediction of transient dynamics. \\

A final aspect that wasn't considered in the current investigation is the testing of higher order nonlinear systems. The simple pendulum problem is considered as a simple non-linear system and may be poorly suited to highlight the benefits of using spectogram image representation of the non-linear state space data. Now that the framework is available, it would be very valuable to test the proposed approach against another non-linear dynamic system like a walking quadruped or biped structure. Additionally, this framework makes it possible that image data could be used in the Koopman operator framework in solving problems outside of control design or system identification.\\

In conclusion, the current investigation showed that the usage of a latent image data representation increases the complexity of the deep learning model, though it doesn't present any increase in prediction accuracy until a lifting dimension of 12. At that dimension, the approach shows a good capture of the transient aspect of the dynamic while the usage of only raw data still shows better prediction over short time. The major advantage that this approach has over previous implementations of DKRC is that it adds the supervised learning component to this problem. This greatly reduces training time and allows for several different heuristics and criteria to be tested against the identified models such that a linear system with desired properties can be found through an external optimization process. \\\\

\bibliography{ref}
\bibliographystyle{IEEEtran}
\end{document}